\documentclass[conference]{IEEEtran}
\usepackage{times}
\usepackage{color}
\usepackage[usenames,dvipsnames,table,xcdraw]{xcolor}

\newcommand{\todo}[1]{}
\renewcommand{\todo}[1]{{\color{red} TODO: {#1}}}
\usepackage[numbers]{natbib}
\usepackage{multicol}
\usepackage{mathtools}
\let\labelindent\relax
\usepackage[shortlabels]{enumitem}
\usepackage{bbm}
\usepackage{bm}
\usepackage{amsmath, amssymb, amscd}
\usepackage{algorithm}
\usepackage[noend]{algpseudocode}
\usepackage[bookmarks=true]{hyperref}
\DeclareMathOperator*{\argmax}{argmax}

\begin{document}

\title{Untangling Dense Non-Planar Knots by Learning Manipulation Features and Recovery Policies} 



\author{
Priya Sundaresan*$^{1}$, Jennifer Grannen*$^{1}$, Brijen Thananjeyan$^{1}$, Ashwin Balakrishna$^{1}$, Jeffrey Ichnowski$^{1}$, \\ Ellen Novoseller$^{1}$,
 Minho Hwang$^1$, Michael Laskey$^{2}$, Joseph E. Gonzalez$^{1}$, Ken Goldberg$^{1}$ \\
}

\makeatletter
\def\thanks#1{\protected@xdef\@thanks{\@thanks \protect\footnotetext{#1}}}
\makeatother
\thanks{
*Equal contribution.}
\thanks{
$^{1}$AUTOLAB at the University of California, Berkeley, USA.}
\thanks{
$^{2}$Toyota Research Institute, Inc., USA.}
\thanks{
Correspondence  to  $\{ \texttt{priya.sundaresan, jenngrannen\}}
\\ \texttt{@berkeley.edu}$}



%

\maketitle

\begin{abstract}
Robot manipulation for untangling 1D deformable structures such as ropes, cables, and wires is challenging due to their infinite dimensional configuration space, complex dynamics, and tendency to self-occlude. 
Analytical controllers often fail in the presence of dense configurations, due to the difficulty of grasping between adjacent cable segments. We present two algorithms that enhance robust cable untangling, LOKI and SPiDERMan, which operate alongside HULK, a high-level planner from prior work. 
LOKI uses a learned model of manipulation features to refine a coarse grasp keypoint prediction to a precise, optimized location and orientation, while SPiDERMan uses a learned model to sense task progress and apply recovery actions.
We evaluate these algorithms in physical cable untangling experiments with 336 knots and over 1500 actions on real cables using the da~Vinci surgical robot. We find that the combination of HULK, LOKI, and SPiDERMan is able to untangle dense overhand, figure-eight, double-overhand, square, bowline, granny, stevedore, and triple-overhand knots. The composition of these methods successfully untangles a cable from a dense initial configuration in 68.3\% of 60 physical experiments and achieves 50\% higher success rates than baselines from prior work.  Supplementary material, code, and videos can be found at \texttt{\url{https://tinyurl.com/rssuntangling}}.

\end{abstract}

\IEEEpeerreviewmaketitle

\section{Introduction}
\label{sec:intro}
Robot manipulation of 1D deformable objects, such as cables, ropes, and wires, can facilitate automation of tasks in industrial, surgical, and household settings~\cite{henrich2012robot}. Such manipulation tasks include organization of cables~\cite{grannen2020untangling,she2019cable}, hoses~\cite{sanchez2020tethered}, suture thread~\cite{van2010superhuman}, and wires used in automotive and other electronic assembly~\cite{jiang2011robotized}, as well as reducing clutter and preventing injury in surgical, manufacturing, and home environments~\cite{cappell2010injury}.
As robots automate more tasks involving 1D deformable objects, which we refer to as ``cables,'' they will increasingly confront highly complex knots, either because the task itself requires knot untangling or because untangling cables is a prerequisite for a downstream task.  Prior work has explored untangling manipulation plans consisting of high-level manipulation primitives (e.g., pick and pull points)~\cite{sundaresan2020learning,lui2013tangled,grannen2020untangling}, but robots often face challenges when attempting to execute these plans in physical manipulation tasks, due to the complexity of dynamics modeling and perception in real-world settings. This motivates the need for a low-level controller which can mitigate the gap between high-level planning and physical execution. This is relevant to any task involving untangling or arrangement of cables where sensing grasping precision or recovering from manipulation failures is critical. We propose a low-level controller consisting of two algorithms that jointly improve the robustness of robot untangling.

\begin{figure}[t!]
  \includegraphics[width=0.48\textwidth]{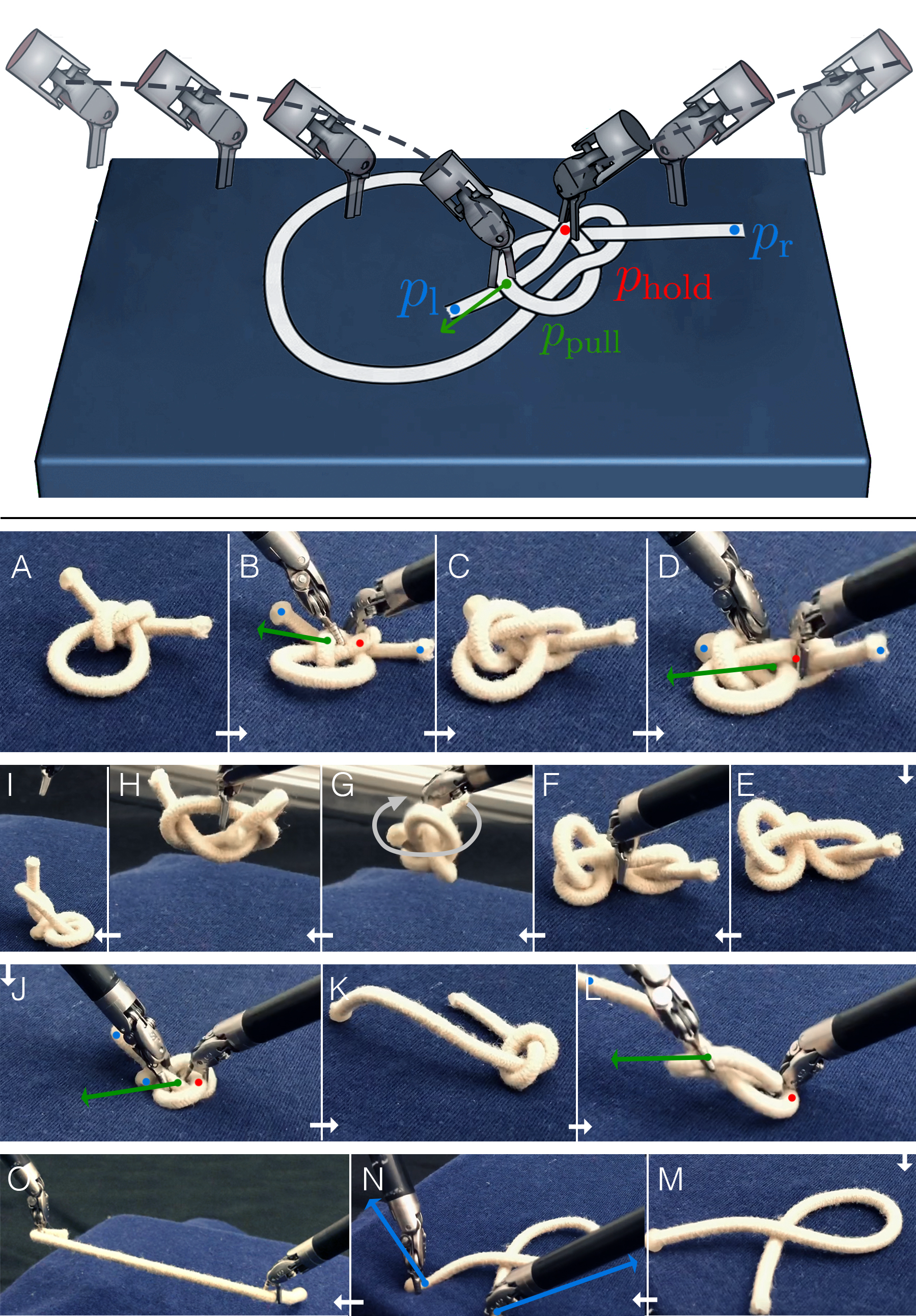}
  \caption{Six-action sequence to untangle a non-planar square knot. We demonstrate the composition of \textbf{HULK} (Hierarchical Untangling from Learned Keypoints), \textbf{LOKI} (Local Oriented Knot Inspection), and \textbf{SPiDERMan} (Sensing Progress in Dense Entanglements for Recovery Manipulation), three algorithms that jointly perform sequential untangling of a dense, non-planar square knot (A). HULK learns to predict task-relevant keypoints for planning bilateral loosening (Node Deletion) and  straightening (Reidemeister) moves. Node Deletion moves sequentially remove a crossing by pulling one cable segment (green) while holding another (red) in place (B-C, D-E, J-K, L-M). LOKI learns manipulation features to perform coarse-to-fine keypoint refinement and infers local cable grasp orientations for fine-grained control. SPiDERMan's recovery policy uses rotation and recentering (Re-Pose) actions that reposition a knot into a new pose at the workspace center for improved grasping (F-I). Finally, a Reidemeister move grasps at predicted endpoints (blue) and pulls taut to yield a linear configuration (O). } 
  \label{splash-fig}
  \vspace{-0.75cm}
\end{figure}

Untangling cables is a challenging task due to the infinite-dimensional configuration space, self-occlusions, and visual homogeneity of cables. These properties complicate the problems of both generating untangling manipulation plans and executing these plans effectively, especially in densely knotted configurations. Prior work for cable manipulation has primarily explored manipulation planning methods that learn full state information~\cite{sundaresan2020learning,lui2013tangled} or task-specific features~\cite{grannen2020untangling}. These methods then execute planned actions through a sequence of analytical, hand-engineered motion primitives, which can be brittle and imprecise under increasing knot complexity and density due to layered self-occlusions, friction at cable intersections, and the  need for precision in grasping. Most prior work considers cable configurations that have sufficiently wide space between self-intersections for a gripper jaw to fit or that have only two cable segments at each self-intersection. These manipulation challenges are exacerbated for \emph{non-planar} configurations, where configuration complexity increases to intersections with three or more segments. 

~\citet{grannen2020untangling} propose HULK, a high-level vision-based planner for untangling of cables in \emph{semi-planar} configurations, where self-intersections involve at most two segments of cable. In this work, we extend HULK with two novel algorithms for non-planar knots.  
The first algorithm, LOKI (Local Oriented Knot Inspection), learns manipulation features to generate high-precision grasping actions. LOKI refines the cable grasp location and orientation from a coarse keypoint by recentering and learning to grasp orthogonally to the cable's axis, as illustrated in Figure~\ref{splash-fig}. The second algorithm, SPiDERMan (Sensing Progress in Dense Entanglements for Recovery Manipulation) detects and recovers from 4 common failure modes: (1) consecutive poor action executions, (2) exceeding a maximum threshold of untangling actions, (3) the cable leaving the reachable workspace, and (4) the robot gripper jaws becoming wedged between two segments of dense cable.
Unlike prior work which focuses on high precision for failure avoidance~\cite{paradis2020intermittent, lee2020guided}, SPiDERMan acknowledges the likelihood of failures in real-world settings and plans recovery actions.  

We perform physical untangling experiments with cables in dense, non-planar configurations more complex than the semiplanar arrangements considered in prior work and find that the combination of HULK, LOKI, and SPiDERMan achieves 50\% higher success rates 
in comparison to analytical controllers.

\section{Related Work}
Deformable object manipulation has seen a surge of interest recently in the robotics research community~\cite{lin2020softgym, yan2020learning, yan2020self, sundaresan2020learning, ganapathi2020learning, hoque2020visuospatial, seita2019deep, seita2020learning}, but modeling state and action spaces in such tasks remains challenging. 
Much prior work focuses on the complex task of generating plans for manipulating cables and higher-dimensional deformable objects. Executing the plans, especially in the presence of failure modes, is often left to future work.  In this work, we focus on addressing untangling-specific failure modes by learning controllers to robustly execute plans for the task.



\subsection{Deformable Object Manipulation}
Several perception-focused techniques have proven effective in multi-step robotic manipulation of cables and other deformable objects such as cloth and bags. One approach performs state estimation and leverages the learned state representation for downstream planning. For instance, \citet{lui2013tangled} and \citet{chi2019occlusion} propose using classical visual feature extraction to estimate the state of deformable rope and cloth, respectively, subject to partial occlusion. \citet{sundaresan2020learning} investigate object representation learning via dense object descriptors~\cite{schmidt2016self, florence2018dense} for rope knot-tying and arrangement, and \citet{ganapathi2020learning} extend this methodology to 2D fabric smoothing and folding. Alternative perception-driven approaches explore learning latent state spaces~\cite{yan2020learning} for cloth manipulation, or using semantic keypoint perception~\cite{grannen2020untangling, yan2020self} for rope tracking. However, robust state estimation remains challenging for heavily self-occluded configurations, as are encountered in densely knotted cables. This work builds upon prior work on keypoint prediction for cable untangling, but unlike previous work, we couple perception with robust learned control policies for grasping and failure recovery. 

Other methods learn visuomotor policies end-to-end. Imitation learning~\cite{seita2019deep}, video prediction models~\cite{hoque2020visuospatial}, model-free reinforcement learning (RL)~\cite{lee2020learning}, and model-based RL~\cite{matas2018sim} have successfully been applied to goal-conditioned fabric manipulation. In other end-to-end approaches, \citet{nair2017combining} learn a dynamics model for performing template shape matching and knot-tying with rope. \citet{zhang2020robots} learn to directly regress robotic configurations in joint space for robotic vaulting of cables over obstacles. While these approaches are broadly general and can flexibly apply to many tasks, they can lack precision and do not leverage geometric structure in the problem, making them difficult to apply to dexterous manipulation tasks such as cable untangling, in which geometric reasoning and fine-grained manipulation are critical.

Several recent works have also explored learning robot policies for deformable manipulation entirely in simulation and transferring them to the real world via domain randomization~\cite{seita2020learning, sundaresan2020learning, yan2020learning, ganapathi2020learning, hoque2020visuospatial, seita2020learning, zhang2020robots}. By contrast, we train perception systems using a combination of data from simulation and the physical world. Crucially, we observe that physical data is critical for tasks that require global reasoning, as it is difficult to simulate the full distribution of configurations encountered during an untangling sequence. However, for tasks that only require local reasoning within specific knots, simulated data can provide sufficient information for robust perception. 

\subsection{Cable Untangling}
To the best of our knowledge, \citet{lui2013tangled} is the first published study of robot cable untying. The authors use RGB-D sensing and classical feature extraction to approximate a tangled cable as a graph consisting of cable crossings and endpoints. We primarily build upon the work of \citet{grannen2020untangling}, which (1) presents a geometric algorithm for cable untangling based on the graphical abstraction from \citet{lui2013tangled} and (2) uses deep learning to detect cable crossings with keypoint inference. These keypoints serve as input to a greedy action planner that iteratively reduces the number of crossings. 

\citet{lui2013tangled} and \citet{grannen2020untangling} present algorithms for generating untangling manipulation plans for cables with semiplanar knots. However, their executed controllers are based on a set of analytical motion primitives, which make simplified assumptions about cable geometry that do not hold in highly deformed or occluded states.

In this work, we focus on the problem of learning a controller to robustly execute the manipulation plans generated by \citet{grannen2020untangling}. First, in adapting the method to a more challenging class of knots, we employ coarse-to-fine refinement strategies to perform robust cable grasping. Several recent works have also studied coarse-to-fine controllers for manipulation tasks requiring precision, including surgical peg transfer~\cite{paradis2020intermittent} and peg insertion~\cite{lee2020guided, johannink2019residual}. Secondly, we explore recovery from manipulation-induced errors in the context of closed-loop untangling. Recent work has also considered recovery from manipulation errors in contact-based robotics tasks~\cite{wang2019learning, donald1989error, recovery-rl}. However, these works do not exploit task geometry, which is particularly relevant to cable untangling. 

\section{Problem Statement}
\label{sec:ps}

This work considers untangling a densely knotted cable from RGB image observations. We use a bilateral robot to hold and pull cable segments until the configuration reaches a fully untangled state with no crossings. The cable color is assumed to be distinguishable from the workspace color. 

We assume access to a bilateral manipulator and overhead RGB image observations from a camera that is a fixed distance above the workspace manipulation surface. We assume the camera is calibrated with a pixel-to-world transform using standard image registration methods. Within the workspace, we assume that the full cable is within the reachable limits of the robot and that the cable width does not exceed the gripper's maximum opening width. These assumptions ensure that both robot arms are able to hold and pull cable segments. We define three points, $\bm{w}_l$, $\bm{w}_c$, and $\bm{w}_r$, which are located respectively on the left side, center, and right side of the workspace for recovery action planning.



We make two assumptions about the initial cable configuration: 
(1) \emph{limited non-planarities}: unlike prior work~\cite{grannen2020untangling} which is limited to 2 segments in each \emph{semi-planar} cable intersection, we allow for up to 3 segments (\emph{non-planar});
and (2) \emph{visible endpoints}: both cable endpoints are unoccluded and visible in overhead RGB images for planning purposes. We distinguish between the two endpoints as left and right based on their pixel coordinate $x$-values, breaking ties arbitrarily. 

We formalize each cable's configuration by extending the graphical cable representation from \citet{lui2013tangled} to non-planar 
configurations. 
Using this modelling approach, we plan untangling actions for an algorithmic supervisor, BRUCE (see overview in Sec.~\ref{sec:bruce}). The physical untangling algorithms---HULK (Sec.~\ref{sec:hulk}), LOKI, and SPiDERMan (Sec.~\ref{sec:methods})---do not explicitly reconstruct this graph, but rather learn relevant features from RGB images to execute actions.


We model cable state via an undirected graph $G = (V, E)$ with cable endpoints and crossing locations represented as vertices $v \in V$ and cable segments between vertices represented as edges $e = (u,v) \in E$ for $u, v \in V$, where $u \neq v$. We use the term ``node'' and ``vertex'' interchangeably and label the nodes corresponding to the left and right endpoints as $v_l$ and $v_r$, respectively. For a node and adjacent edge $(v, e) \in (V, E)$, we annotate the graph with $X(v, e) \in \{+1, -1, -2\}$ (Equation~\ref{eq:annotations}) to designate the cable segment hierarchy at each node:
\begin{align}
\hspace{-0.2cm}
\label{eq:annotations}
X(v, e) = 
\begin{cases} 
+1 & \parbox{15em}{if $v$ is an endpoint or if $e$ crosses above all other edges at $v$} \\ -1 & \text{if $e$ crosses below only one edge at $v$} \\ -2 & \text{if $e$ crosses below two other edges at $v$.} 
\end{cases}
\end{align}

\begin{figure}[t!]
\hspace{-0.5cm}
  \includegraphics[width=0.49\textwidth]{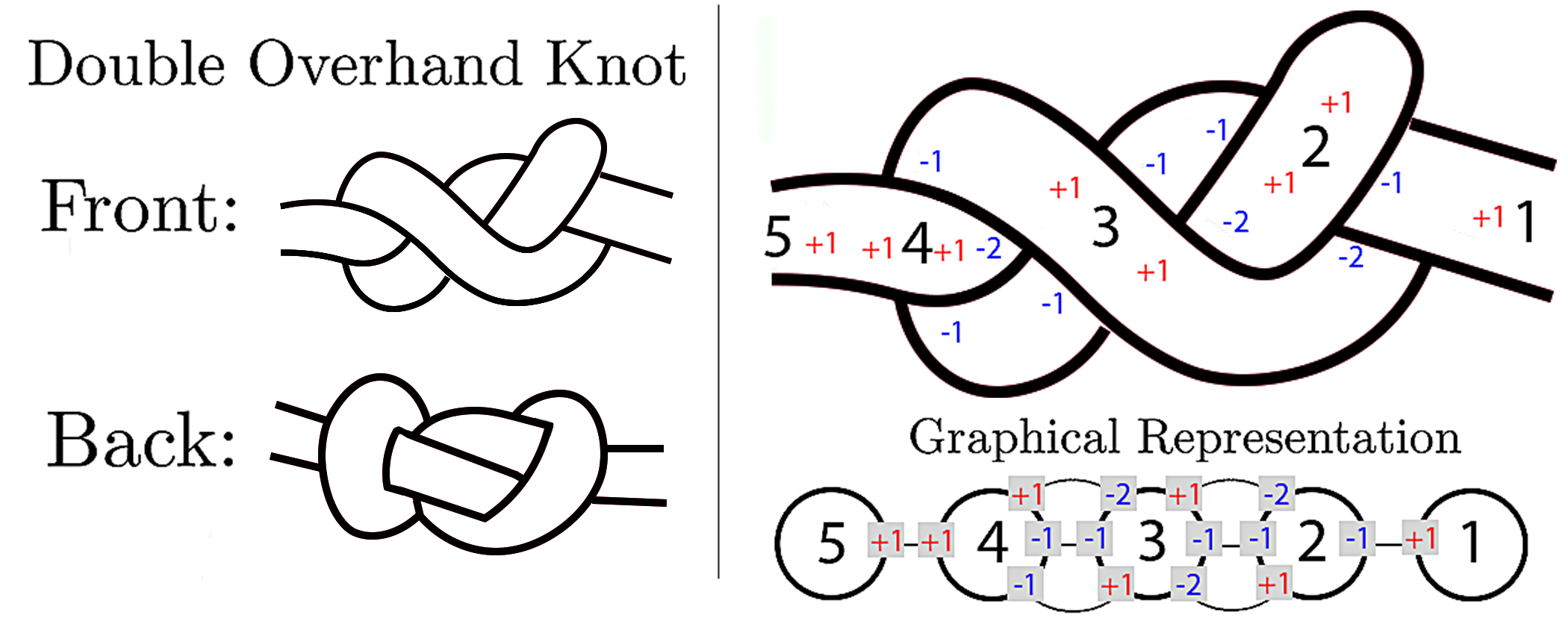}
  \vspace{-0.4cm}
  \caption{\textbf{Non-Planar Graphical Representation:} We model cable configuration with a graph as in \citet{lui2013tangled} and extend this representation to non-planar configurations with an updated annotation method in Equation~\eqref{eq:annotations}. We visualize a dense double overhand knot from two views to illustrate non-planarity, along with its corresponding annotated graph.} 
  \label{ps-fig}
  \vspace{-0.65cm}
\end{figure}

We define an \emph{under-crossing} as a set of one node $v$ and two incident edges $e_i, e_j \in E$, where $e_i = (v, v')$ and $e_j = (v, v'')$ for some $v', v'' \in V$, $X(e_i, v) = X(e_j, v) \in \{-1, -2\}$, and $e_i, e_j$ are contiguous in the cable. Thus, the cable segment represented by $e_i, e_j$ is occluded by one or two other cable segments. Similarly, an \emph{over-crossing} is a set of one node $v$ and two incident edges $e_i = (v, v')$ and $e_j = (v, v'')$ for some $v', v'' \in V$, where $X(e_i, v) = X(e_j, v) = +1$ and $e_i, e_j$ are contiguous in the cable.
Under the non-planar assumption, each node has a degree of at most 6, corresponding to 3 cable segments at a crossing. Edges with the same annotation $X(v, e)$ at a node $v$ represent the same cable segment on either side of the crossing. A cable with $|V| = 2$ has only endpoint nodes and no crossings, and thus is fully untangled.

Each gripper performs either a holding, pulling, or rotating action at each time $t$. Below, we define all actions with respect to the 2D image frame. The left robot arm performs a pulling action, $\bm{a}_{t,l}$, by grasping the cable at the pixel $(x_{t,l}, y_{t,l})$ with a (vertical) $z$-axis grasping rotation of $\theta_{t,l}$, lifting by a fixed amount, pulling to $(x_{t,l} + \Delta x_{t,l}, y_{t,l} + \Delta y_{t,l})$, and releasing the cable. Holding actions are indicated via $(\Delta x_{t,l}, \Delta y_{t,l}) = (0,0)$. We denote moving between points without grasping the cable via $\mathbbm{1}_\mathrm{grasp} = 0$. Lastly, $\Delta \theta_{t,l}$ (in degrees) indicates rotating a grasped cable about the $z$-axis. The right robot arm uses an analogous action formulation, $\bm{a}_{t,r}$. We show the full action representation for both arms in Equation~\eqref{eq:actions}:

\begin{flalign}
\label{eq:actions}
\begin{split}
\bm{a}_{t,l} &= (x_{t,l}, y_{t,l}, \theta_{t,l}, \Delta x_{t,l}, \Delta y_{t,l}, \Delta \theta_{t,l}, \mathbbm{1}_\mathrm{grasp})\\
\bm{a}_{t,r} &= (x_{t,r}, y_{t,r}, \theta_{t,r}, \Delta x_{t,r}, \Delta y_{t,r}, \Delta \theta_{t,r}, \mathbbm{1}_\mathrm{grasp}).
\end{split}
\end{flalign}
This action space generalizes that of \citet{grannen2020untangling}, which only parameterizes grasp points and pull vectors.

\section{Preliminaries}
In this section, we review the BRUCE and HULK algorithms introduced in \citet{grannen2020untangling} and discuss relevant algorithmic modifications to accommodate non-planar knots. 

\subsection{BRUCE: Basic \hspace{-0.13cm} Reduction \hspace{-0.13cm} of Under-Crossing \hspace{-0.13cm} Entanglements}
\label{sec:bruce}
\citet{grannen2020untangling} study semi-planar knot untangling given observations of the knot's graphical representation, and propose Basic Reduction of Under-Crossing Entanglements (BRUCE), an algorithm to iteratively untangle knots starting from the rightmost endpoint. 
BRUCE makes use of two manipulation primitives adapted from \citet{lui2013tangled}: (1) \textbf{Reidemeister moves} pull the cable endpoints apart to reduce self-occlusions that are not part of a knot, and (2) \textbf{Node Deletion moves} delete a crossing in the graph by holding an over-crossing edge in place with one arm while the other arm pulls out the cable from the corresponding under-crossing.

BRUCE takes a Reidemeister move to disambiguate the cable state, followed by alternating Node Deletion and Reidemeister moves to sequentially delete crossings and attempt to linearize the cable. In semi-planar configurations, Node Deletion moves pull out the under-crossing edge (labeled $-1$) exiting the first node $v$ traced from the right endpoint, while holding a corresponding over-crossing edge $(+1)$ at $v$ in place.

This work extends Node Deletion moves to operate at non-planar crossings, which contain under-crossing edges labeled as both $-1$ and $-2$. In this setting, a Node Deletion move holds an over-crossing edge (+1) and pulls out the under-crossing edge ($-1$ or $-2$) traced from the right endpoint. This action reduces $|E|$ and $|V|$ by at least 2 and 1, respectively, until the algorithm terminates with $|V| = 2$: a fully linear cable with no crossings and two endpoints.

\subsection{HULK: Hierarchical Untangling from Learned Keypoints}
\label{sec:hulk}
\citet{grannen2020untangling} consider cable untangling from RGB image observations, where the ground truth cable graph representation is not directly observable. To infer untangling actions while bypassing the need for full graph reconstruction, \citet{grannen2020untangling} propose Hierarchical Untangling from Learned Keypoints (HULK), which predicts keypoints from images to execute BRUCE's manipulation primitives. HULK learns a mapping from an image to four Gaussian heatmaps centered at task-relevant keypoints, $f: \mathbb{R}^{640 \times 480  \times 3} \mapsto \mathbb{R}^{640 \times 480 \times 4}$. The means of the four heatmap outputs, denoted by $\bm{\hat{p}}_l,  \bm{\hat{p}}_r, \bm{\hat{p}}_\mathrm{pull},$ and $\bm{\hat{p}}_\mathrm{hold}$, indicate the predicted pixel locations of the left and right cable endpoints ($\bm{\hat{p}}_l$ and $\bm{\hat{p}}_r$) and the predicted pull and hold grasp locations for the next planned Node Deletion move ($\bm{\hat{p}}_\mathrm{pull}$ and $\bm{\hat{p}}_\mathrm{hold}$). 
HULK learns to predict the pull and hold keypoints at the first under/over-crossing pair relative to the rightmost endpoint.

In this work, we adapt the HULK training procedure from \citet{grannen2020untangling}, originally trained on semiplanar knots, to operate on dense, non-planar configurations.
Since training HULK only requires light supervision in the form of keypoints, we follow the training procedure in~\cite{grannen2020untangling} and train HULK on 200 pairs of images and hand-annotations for non-planar configurations. As in \cite{grannen2020untangling}, we use a Gaussian standard deviation of 8 px for the heatmap training data centered about each hand-labelled keypoint, and we apply various data augmentation techniques including image adjustments and affine transformations to generate a training dataset of 3,500 examples. Once trained, HULK uses the predicted keypoints $\bm{\hat{p}}_l,  \bm{\hat{p}}_r, \bm{\hat{p}}_\mathrm{pull},$ and $\bm{\hat{p}}_\mathrm{hold}$ to plan a bilateral move $(\bm{a}_{t,l}, \bm{a}_{t,r})$ at time $t$ for the left and right arms, respectively, as follows.

For a \textbf{Node Deletion} move, the right arm holds the configuration at $\bm{\hat{p}}_\mathrm{hold}$ while the left arm grasps at $\bm{\hat{p}}_\mathrm{pull}$ and pulls in the direction away from $\bm{\hat{p}}_\mathrm{hold}$ by some fixed vector $\bm{n} = (n_x, n_y)$ to slacken the cable. Both arms use coarse analytical grasp rotations $\hat{\theta}_{\text{pull}} = \arctan \left({\frac{\hat{p}_{\text{pull},y} - \hat{p}_{\text{hold},y}}{\hat{p}_{\text{pull},x} - \hat{p}_{\text{hold},x}} }\right)$ 
and $\hat{\theta}_{\text{hold}} = \hat{\theta}_{\text{pull}} + 90^{\circ}$:
\begin{flalign*}
\bm{a}_{t,l} &=  ( \hat{p}_{\text{pull},x}, \hat{p}_{\text{pull},y},
\hat{\theta}_{\text{pull}},
n_x, n_y
, 0, 1) \\ \bm{a}_{t,r} &= ( \hat{p}_{\text{hold},x}, \hat{p}_{\text{hold},y}, \hat{\theta}_{\text{hold}}, 0, 0, 0, 1). 
\end{flalign*}

For a \textbf{Reidemeister} move, the left and right arms simultaneously grasp at $\bm{\hat{p}}_l$ and $\bm{\hat{p}}_r$ and pull the endpoints to opposing workspace ends $\bm{w}_l$ and $\bm{w}_r$, with analytical grasp rotations $\hat{\theta}_{l}$ and $\hat{\theta}_{r}$ obtained as the angle between $(1,0)$ and the vector orthogonal to the first principal component of a masked crop around $\bm{\hat{p}}_{l}$ and $\bm{\hat{p}}_{r}$:

\begin{flalign*}
\bm{a}_{t,l} &= ( \hat{p}_{l,x}, \hat{p}_{l,y}, \hat{\theta}_{l},  w_{l,x} - \hat{p}_{l,x}, w_{l,y} - \hat{p}_{l,y}, 0, 1 ) \\ \bm{a}_{t,r} &= ( \hat{p}_{r,x}, \hat{p}_{r,y}, \hat{\theta}_{r}, w_{r,x} - \hat{p}_{r,x}, w_{r,y} - \hat{p}_{r,y}, 0, 1).
\end{flalign*}

\begin{figure}[t!]
\centering
  \includegraphics[width=0.48\textwidth]{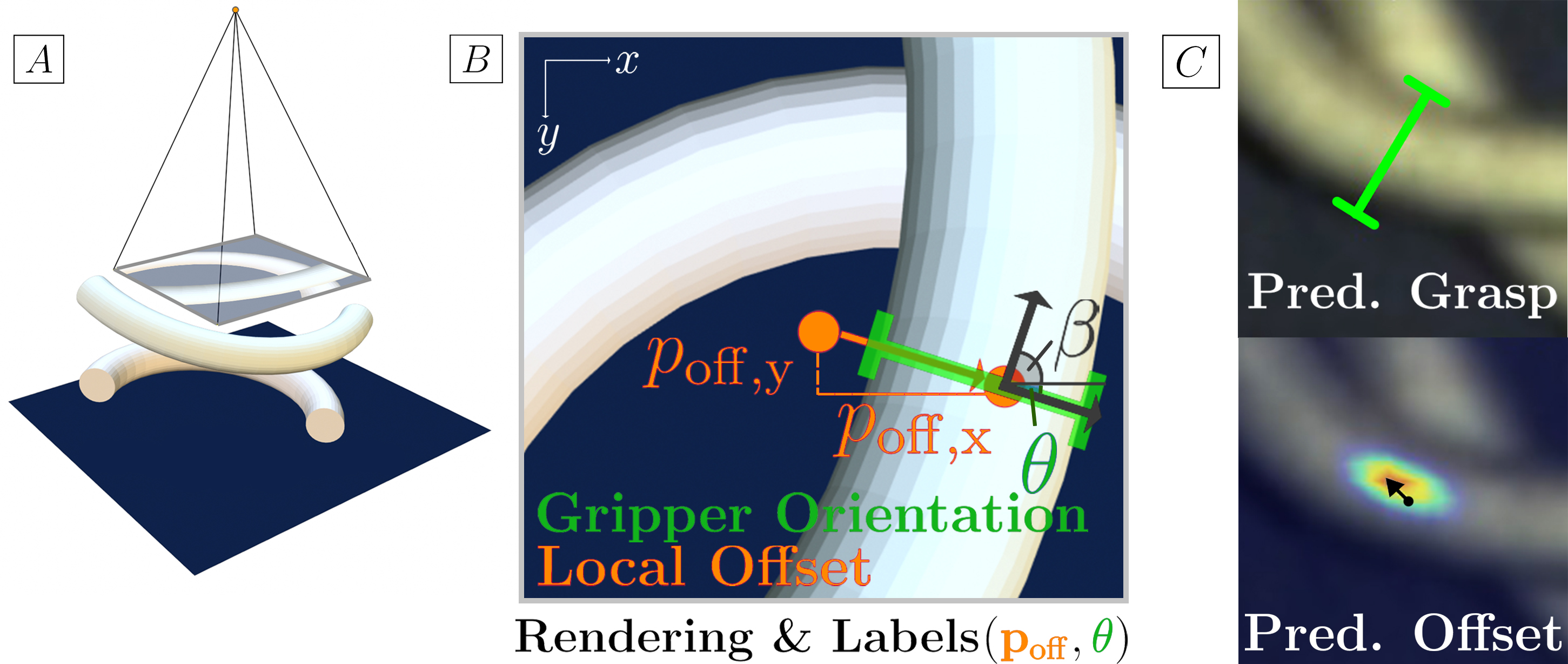}
  \caption{\textbf{LOKI}: LOKI learns precise cable grasping from simulation. We simulate cable crossings in Blender 2.80 as overlapping, warped cylindrical meshes (A) and render RGB crops (B). Each $200\times200$ pixel crop is annotated with a ground truth optimal grasp angle $\theta$ orthogonal to the topmost cylinder orientation $\beta$ and a local offset $(p_\mathrm{off,x}, p_\mathrm{off,y})$ from a coarse grasp prediction to recenter the grasp location along the cable width. At test time (C), LOKI takes a real cable crop centered at a HULK keypoint, refines the keypoint via recentering, and estimates a robust gripper orientation for grasping.
  } 
  \label{loki-fig}
  \vspace{-0.5cm}
\end{figure}

\section{Methods}
\label{sec:methods}
We present two new algorithms, LOKI (Local Oriented Knot Inspection) and SPiDERMan (Sensing Progress in Dense Entanglements for Recovery Manipulation), which are combined to learn a robust controller to increase robustness of high-level manipulation plans from~\citet{grannen2020untangling}.

\subsection{LOKI: Local Oriented Knot Inspection}
To enable finer-grained grasp planning for the knot untying primitives of HULK, we introduce a low-level controller called LOKI. This controller jointly infers robust antipodal grasps that enclose a cable segment orthogonally and performs coarse-to-fine refinement of keypoints. These adjustments are designed to prevent near-miss grasps, a common failure mode in HULK~\cite{grannen2020untangling}. While prior work in cable manipulation has employed analytical grasp planning \cite{sundaresan2020learning, grannen2020untangling}, such heuristic methods fail to generalize to dense, non-planar configurations. Thus, we propose LOKI: Local Oriented Knot Inspection. LOKI maps a locally-cropped cable image, centered at one of the four HULK keypoints $\bm{\hat{p}} \in \{\bm{\hat{p}}_l, \bm{\hat{p}}_r, \bm{\hat{p}}_\mathrm{pull}, \bm{\hat{p}}_\mathrm{hold}\}$, to (1) $\theta$: an angle about the $z$-axis denoting the top-down grasp orientation and (2) $\bm{p}_\mathrm{off} = (p_{\mathrm{off},x}, p_{\mathrm{off},y})$: a local offset in pixel space to recenter the keypoint along the cable width. We implement LOKI as a multi-headed deep neural network with a ResNet-18 \cite{he2016identity} backbone that learns a mapping $g: \mathbb{R}^{200 \times 200 \times 3} \mapsto ([0, 180], \mathbb{R}^{200 \times 200})$. These outputs correspond to the predicted $z$-axis rotation (in degrees) and an unnormalized 2D heatmap centered at the refined keypoint $\bm{\hat{p}} + \bm{\hat{p}}_{\mathrm{off}}$. We discuss the procedure for obtaining the heatmap and rotation training labels below.


\subsubsection{Dataset Generation} We train LOKI in a self-supervised fashion from synthetic images. Simulation provides two advantages in this setting: (1) LOKI requires reasoning about local cable self-intersections which can be approximated with geometric models more readily than global knotted configurations, and (2) hand-annotation of the cable orientation is tedious to obtain from real data but readily accessible from synthetic meshes. We implement a simulation environment in Blender~2.80~\cite{lallemand1998blender} that models cable self-intersections as two overlapping cylindrical meshes slightly warped to emulate the curvature of real cables (see Fig.~\ref{loki-fig}). We generate varied training data by positioning a synthetic overhead camera above, but not centered on, the topmost cylinder. This cylinder is parameterized by a pitch angle $\beta$ denoting its orientation in the $xy$-plane and a translation projected to pixel coordinates $(u,v)$. For each synthetic image rendered, we record the desired gripper orientation as $90^{\circ} + \beta$ to model a grasp enclosing the cable orthogonally and a ground-truth Gaussian heatmap $\mathcal{N}((u, v), \sigma^2 \mathcal{I})$ over the network input image centered at the desired grasp location $(u,v)$, where $\mathcal{I}$ is a $2 \times 2$ identity matrix. Empirically, we find that $\sigma=5\text{~px}$ yields stable training without producing high-entropy heatmaps. LOKI is trained from 3,500 such examples using mean-squared error in degrees for the orientation output's loss and binary cross entropy loss for the heatmaps. 

Given a global image $I \in \mathbb{R}^{640 \times 480 \times 3}$ and a HULK keypoint $\bm{\hat{p}}$, we take a $60 \times 60$ crop centered at $\bm{\hat{p}}$. We resize this crop to the LOKI network input dimensions, $200 \times 200$, resulting in a cropped image $\hat{I}$. LOKI yields outputs $[\hat{\theta}, H] = g(\hat{I})$, such that $\bm{\hat{p}}_{\mathrm{off}}$ is given by the highest-probability point in $H$: 
\[(\hat{p}_{\mathrm{off},x}, \hat{p}_{\mathrm{off},y}) = \gamma \left[\left(\argmax_{(u_i,v_i) \in \hat{I}} H[u_i, v_i]\right) - (u_c,v_c)\right],\]
where $\gamma = 60/200$ is a downscaling factor to account for crop resizing, and $(u_c, v_c) = (100,100)$ is the upscaled crop center (i.e., a coarse keypoint prediction).
The refined keypoint $(\hat{p}_x + \hat{p}_{\mathrm{off},x}, \hat{p}_y + \hat{p}_{\mathrm{off},y})$ and predicted gripper orientation $\hat{\theta}$ are used for planning grasps across all manipulation primitives.

\subsection{SPiDERMan: Sensing Progress in Dense Entanglements for Recovery Manipulation}
To sense and recover from manipulation failures, we propose SPiDERMan. SPiDERMan uses both learned and analytical methods to sense untangling progress and employs two novel manipulation primitives to avoid and recover from failures. We address four manipulation failure modes from~\citet{grannen2020untangling}: (1) consecutive poor action executions due to high cable density, (2) task incomplete after exceeding the maximum number of untangling actions, (3) the cable leaving the workspace during manipulation, and (4) the high-density cable becoming wedged in the robot jaws. 

We first discuss how SPiDERMan detects each failure mode from RGB image inputs and next define two manipulation primitives for failure mode recovery.

\subsubsection{Detection}
\label{sec:methods-perception}

SPiDERMan employs one learning-based and two analytical perception methods to detect or prevent each of the four failure modes. To address the first two failures, SPiDERMan compares workspace image observations $I_{t'} \in \mathbb{R}^{640 \times 480 \times 3}$ before and after executing actions, where $t'$ denotes the number of untangling, or Node Deletion, actions executed thus far. SPiDERMan is implemented with a ResNet-18 backbone and learns a classifier $d : \mathbb{R}^{640 \times 480 \times 6} \mapsto \{0,1\}$, trained with a binary cross-entropy loss.
To determine when a recovery action is needed due to lack of untangling progress at test time, $d(I_1, I_2)$ uses a simple decision threshold of 0.5 over the softmax of the classifier output to determine whether image $I_1$ corresponds to a denser (1) or less dense (0) state than $I_2$. When two successive actions have not effectively loosened the knot, the following condition evaluates to true:
\begin{align}
\label{eq:rotate_condition}
d(I_{t'}, I_{t'-1}) \text{ and } d(I_{t'-1}, I_{t'-2}),
\end{align}
and we conclude that the configuration is pathological and a recovery action will be performed.
In this way, SPiDERMan recognizes untangling progress over \emph{multiple} actions, removing the need for a fixed untangling action limit. Instead, we define a termination condition:
\begin{align}
\label{eq:termination}
d(I_{t'}, I_{t'-5}) \text{ or } d(I_{\text{ref}}, I_{t'})
\end{align}
that checks both for untangling success against an untangled reference cable image $I_{\text{ref}}$ and for progress every 5 actions to prevent termination as long as there is progress.  Condition \eqref{eq:termination} evaluates to true when no progress has occurred over the last 5 actions, or if the cable reaches a fully-untangled state.

\begin{figure*}[htbp!]
\centering
  \includegraphics[width=0.99\textwidth]{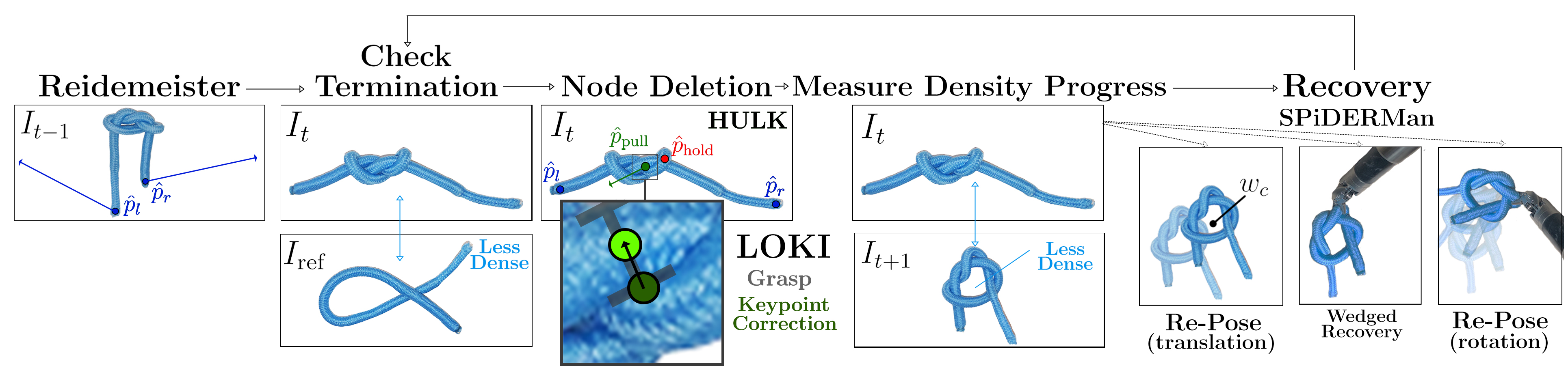}
  \caption{\textbf{System Overview}: We illustrate Algorithm~\ref{alg:untangle_alg} used to perform the cable untangling shown in Figure~\ref{splash-fig}. Starting from the left, we untangle a cable from a non-planar initial configuration following the actions outlined by BRUCE: one initial Reidemeister move followed by successive Node Deletion moves until no crossings remain. HULK instantiates BRUCE's Reidemeister and Node Deletion primitives by regressing 4 keypoints (left endpoint, pull, hold, and right endpoint) from RGB image inputs to define high-level action plans. 
  LOKI refines each action by predicting a local offset to center each keypoint along the cable width and a gripper orientation rotation to grasp orthogonally to the cable direction. SPiDERMan senses action success by comparing configuration density in image observations before and after a Node Deletion action is performed. SPiDERMan also employs contour-detection methods (not shown) for sensing when the cable is approaching workspace limits or when the cable is wedged in the gripper jaws. When a lack of progress or a failure mode is detected, SPiDERMan performs one of two failure recovery manipulation primitives: a Wedged Recovery move or a Re-Posing move (Rotation or Translation). It checks when a cable is fully untangled by comparing its density to that of a pre-defined reference image $I_{\text{ref}}$ with a single crossing.
  } 
  \label{system-overview-fig}
  \vspace{-0.68cm}
\end{figure*}

When $I_{t'}$ is fully untangled, both $I_\mathrm{ref}$ and $I_{t'}$ are equally dense. For this reason, we use an image observation of a knot-free cable with a single crossing for $I_\mathrm{ref}$ (Figure~\ref{system-overview-fig}) to prevent termination false negatives arising from a bias toward $I_\mathrm{ref}$. 

SPiDERMan also applies analytical contour-based perception to implement recovery manipulation primitives. Given a workspace image observation $I_t$, we preprocess the image by converting to grayscale and applying binary thresholding. The algorithm detects all contiguous contours in the resulting image using the open-source implementation from \citet{suzuki1985topological} in OpenCV~\cite{bradski2008learning}. We extract an approximate contour of the cable $\mathcal{P} = \{\bm{p}_1, \bm{p}_2, \hdots, \bm{p}_n\}$ via contour area filtering. Next, SPiDERMan approximates the cable center as the mean point $\bm{\bar{p}}$ of the contour, projected to the lie on the cable: $\bm{\hat{p}}_\mathrm{c} = \arg \min_{\bm{p}_i \in \mathcal{P}} || \bm{p}_i - \bm{\bar{p}}||$. Between actions, the left and right grippers move to known poses with gripper jaw pixel coordinates $\bm{g}_l$ and $\bm{g}_r$, respectively, outside of the field of view of the untangling workspace. Given $\bm{\hat{p}}_\mathrm{c}$, $\bm{g}_l$, and $\bm{g}_r$, Condition \eqref{eq:gripper_stuck} detects when the cable is wedged in either of the grippers by checking when the distance between the approximate cable center, $\bm{\hat{p}}_\mathrm{c}$, and either gripper tooltip, $\bm{g}_{r}$ or $\bm{g}_l$, is below a hand-tuned threshold of 20~px: 
\begin{align}
\label{eq:gripper_stuck}
\min\{||\bm{\hat{p}}_\mathrm{c} - \bm{g}_{l}||, ||\bm{\hat{p}}_\mathrm{c} - \bm{g}_{r}||\} < 20\text{~px.}
\end{align}

To detect when the cable is leaving the workspace, we define a second condition over the contour-based perception method above to check when the center of the cable mask $\bm{\hat{p}}_\mathrm{c}$ exceeds a hand-tuned 200~px threshold from the workspace center $\bm{w}_c$:
\begin{align}
\label{eq:leaving_workspace}
||\bm{\hat{p}}_\mathrm{c} - \bm{w}_c|| > 200\text{~px.}
\end{align}


\subsubsection{Recovery}
We define two novel manipulation primitives to recover once a failure is detected: (1) \textbf{Re-Posing} moves reorient and place a cable at the workspace center, and (2) \textbf{Wedged Recovery} moves detach a gripper jaw wedged between two cable segments.

For a \textbf{Re-Posing} move, we improve the Recentering primitive from~\citet{grannen2020untangling} to rotate the cable in addition to centering the configuration in the workspace. The right arm first grasps the cable at the center of the configuration $\bm{\hat{p}}_c$ with a grasp location $\bm{\hat{p}}_c + \bm{\hat{p}}_{c, \text{off}}$ and grasp rotation of $\hat{\theta}_c$ predicted by LOKI. The right arm then lifts by a fixed amount and moves the cable to the predefined workspace center $\bm{w}_c$.
When two successive poor actions are detected via Equation \eqref{eq:rotate_condition}, the right gripper then rotates by $180^{\circ}$ to re-pose the configuration before releasing the grasp. A \emph{Re-Posing move (Rotation)} includes this rotation, while a \emph{Re-Posing move (Translation)} does not:

\begin{flalign*}
\bm{a}_{t,r} = ( \hat{p}_{c,x}, \hat{p}_{c,y}, \hat{\theta}_c, w_{c,x} - \hat{p}_{c,x}, w_{c,y} - \hat{p}_{c,y}, \underbrace{\{0^{\circ}, 180^{\circ}\}}_{\text{optional rotation}}, 1 ).
\end{flalign*}

A \textbf{Wedged Recovery} move is comprised of two successive actions. When detecting that the left robot gripper jaw is wedged between cable segments, the left arm first brings the stuck cable to the workspace center $\bm{w}_c$ and releases its grasp. Given the detected cable contour $\mathcal{P}$, the right arm grasps the cable at the far right point on the cable mask $\bm{\hat{p}}_{\text{hold}} = \arg \min_{\bm{p}_i \in \mathcal{P}} || \bm{p}_i - \bm{w}_r||$ with a grasp location $\bm{\hat{p}_{\text{hold}}} + \bm{\hat{p}_{\text{hold},\text{off}}}$ and grasp rotation $\hat{\theta}_{\text{hold}}$ given by LOKI and holds it down on the workspace while the left arm returns to its home position, a fixed height above the predefined point $\bm{w}_l$:
\begin{flalign*}
\bm{a}_{t,l} &= ( w_{l,x}, w_{l,y}, 0, w_{c,x}-w_{l,x}, w_{c,y}-w_{l,y}, 0, 1 ) \\ \bm{a}_{t+1,r} &= ( \hat{p}_{\text{hold},x}, \hat{p}_{\text{hold},y}, \hat{\theta}_{\text{hold}}, 0, 0, 0, 1 ) \\ \bm{a}_{t+1,l} &= ( w_{c,x}, w_{c,y}, 0, w_{l,x}-w_{c,x}, w_{l,y}-w_{c,y}, 0, 0 ).
\end{flalign*}

When the cable is wedged in the right robot gripper, we execute the equivalent procedure with the roles of the right and left arms switched. In this case, $\bm{\hat{p}}_{\text{hold}}$ represents the far left point on the cable mask, and the right arm returns to the home position above the predefined point $\bm{w}_r$.

Algorithm \ref{alg:untangle_alg} presents the untangling algorithm combining HULK from~\citet{grannen2020untangling} with LOKI and SPiDERMan, and Figure~\ref{system-overview-fig} illustrates the system.

\begin{algorithm}
\caption{Untangling with LOKI and SPiDERMan}
\begin{algorithmic}
\State \textbf{Input:} RGB image of cable

\State Reidemeister move with $\underbrace{\bm{\hat{p}}_r}_{\text{HULK}} + \underbrace{\bm{\hat{p}}_{r,\text{off}}, \hat{\theta}_{r}}_{\text{LOKI}}$, $\bm{\hat{p}}_l + \bm{\hat{p}}_{l,\text{off}}$, $\hat{\theta}_{l}$
\While{NOT Eq. \eqref{eq:termination}} \Comment{Termination check}
\While{Eq. \eqref{eq:gripper_stuck}} \Comment{Wedge check}
\State 
{\small
Wedge Recovery with $\bm{\hat{p}}_{\text{hold}} + \bm{\hat{p}}_{\text{hold}, \text{off}}$, $\hat{\theta}_{\text{hold}}$, $\bm{w}_c$, $\bm{w}_{l/r}$
}
\EndWhile
\If{Eq. \eqref{eq:rotate_condition}} \Comment{Poor actions check}
\State 
{\small
Re-Posing (Rotation) with $\bm{\hat{p}}_c + \bm{\hat{p}}_{c, \text{off}}$, $\hat{\theta}_c$, $\bm{w}_c$}
\EndIf
\If{Eq. \eqref{eq:leaving_workspace}} \Comment{Leaving workspace check}
\State 
{\small
Re-Posing (Translation) with $\bm{\hat{p}}_c + \bm{\hat{p}}_{c, \text{off}}$, $\hat{\theta}_c$, $\bm{w}_c$}

\EndIf
\State 
{\small Node Deletion with $\bm{\hat{p}}_{\text{hold}} + \bm{\hat{p}}_{\text{hold}, \text{off}}$, $\hat{\theta}_{\text{hold}}$, $\bm{\hat{p}}_{\text{pull}} + \bm{\hat{p}}_{\text{pull}, \text{off}}$, $\hat{\theta}_{\text{pull}}$}

\EndWhile
\State \textbf{return} DONE

\end{algorithmic}
\label{alg:untangle_alg} 
\end{algorithm}

\section{Experiments}
We experimentally evaluate the separate and joint effectiveness of HULK, LOKI, and SPiDERMan in performing physical untangling of dense semi-planar and non-planar knots. 

\subsection{Overview of Policies}
We compare (1) the proposed method (Alg.~\ref{alg:untangle_alg}) synthesizing HULK, LOKI, and SPiDERMan (H+L+S) against three baselines: (2) HULK alone (H), (3) HULK with only LOKI (H+L), and (4) HULK with only SPiDERMan (H+S). Notably, baselines (2) and (4) employ analytical grasp planning~\cite{grannen2020untangling}. In this strategy, Reidemeister grasps are planned orthogonally to the major axis given by principal component analysis on masked endpoint crops. Node Deletion grasps are also heuristically planned and are orthogonal to predicted action vectors as described in Sec.~\ref{sec:hulk}. Baselines (2) and (3) do not use any recovery manipulation primitives. All policies employ the termination condition given by Eq.~\eqref{eq:termination}.

We train HULK and SPiDERMan on the same dataset of 320 cropped, $640 \times 480$ real RGB workspace image observations augmented 8X and centered around detected cable contours according to Section~\ref{sec:methods-perception}. We also train LOKI from 3,000 $200 \times 200$ synthetically-generated cable crops.

\subsection{Tiers of Difficulty}
\begin{figure}[t!]
  \includegraphics[width=0.49\textwidth]{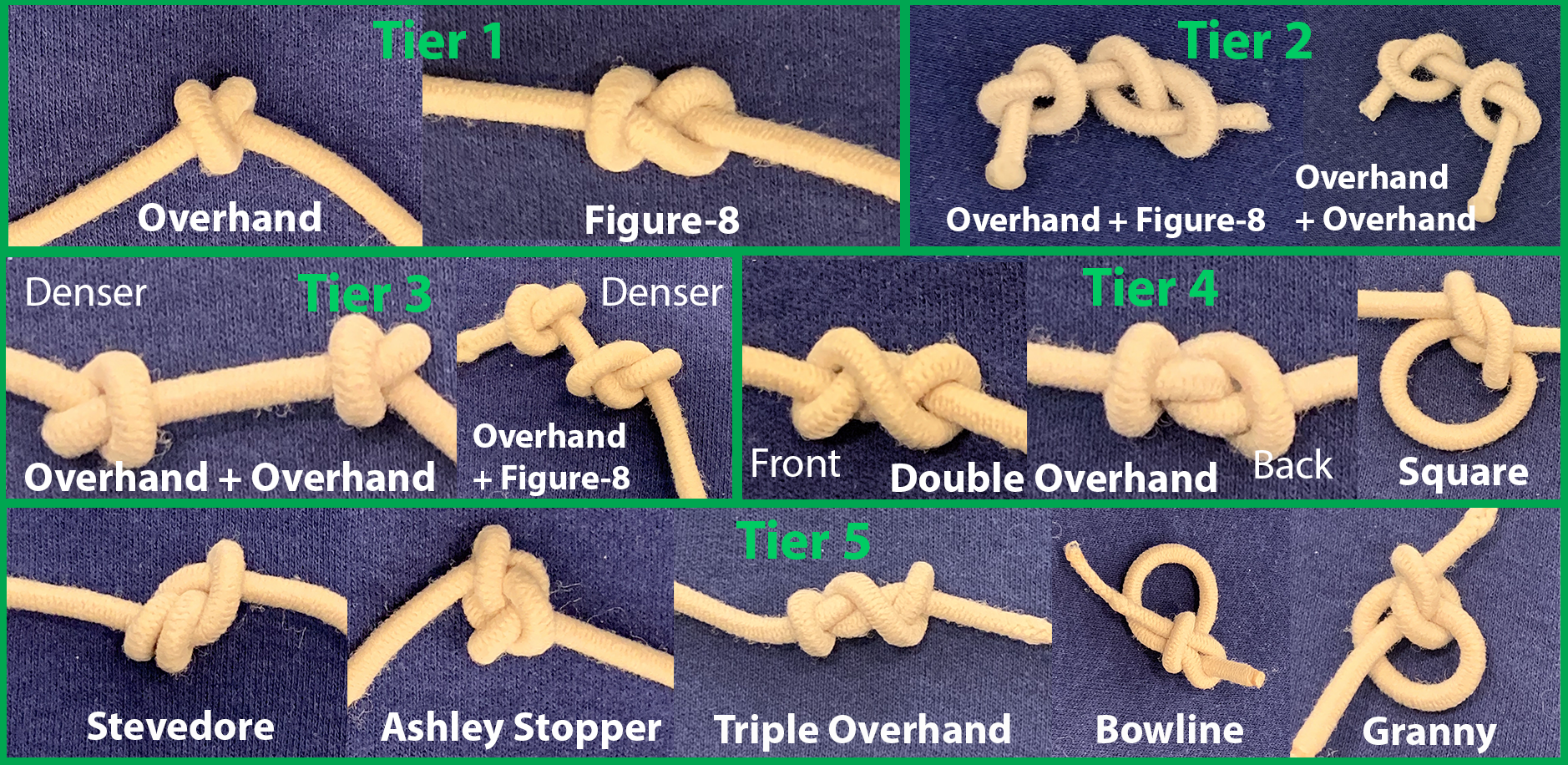}
  \vspace{-0.6cm}
  \caption{\textbf{Difficulty Tiers:} Examples of dense initial configurations used in physical untangling experiments. We categorize configurations into 5 tiers of difficulty based on configuration complexity (semi-planar vs. non-planar), knot density (qualitatively measured), number of knots, and whether these knots were present in the training data. Tiers 1 and 2 are semi-planar and are drawn from~\cite{grannen2020untangling}, while Tier 3 contains semi-planar configurations denser than those in~\cite{grannen2020untangling}, and Tiers 4 and 5 consist of non-planar knots.
  } 
  \label{configs-fig}
  \vspace{-0.6cm}
\end{figure}

All policies are tested on \emph{dense} knots across five difficulty tiers (Fig.~\ref{configs-fig}), categorized by the degree of non-planarity and whether the knot class appeared during training:
{ \footnotesize
\begin{description}
    \item \textbf{Tier 1}: A single semiplanar knot (figure-eight or overhand); knots of this class were present in training data, as in 
    \cite{grannen2020untangling}
    \item \textbf{Tier 2}: Two semiplanar knots (figure-eight or overhand); knots of this class were present in training data, as in 
    \cite{grannen2020untangling}
    \item \textbf{Tier 3}: Two semiplanar knots (figure-eight or overhand); knots of this class were present in training data and denser than those of \cite{grannen2020untangling} 
    \item \textbf{Tier 4}: A single non-planar knot (double overhand or square); knots of this class were present in training data
    \item \textbf{Tier 5}: A single non-planar knot (stevedore, bowline, ashley stopper, granny, or heaving line); knots of this class were \emph{not} present in training data
\end{description}
}

HULK, LOKI, and SPiDERMan were trained on configurations containing up to two overhand and figure-eight knots, single double overhand knots, and single square knots. While Tiers 1-4 contain knots types that appear in the training dataset, Tier 5 tests the generalization capabilities of HULK, LOKI, and SPiDERMan to knot types absent in the training data.

\subsection{Experimental Setup}
We execute all experiments using the bilateral dVRK robot ~\cite{kazanzides-chen-etal-icra-2014} equipped with two 7-DoF arms. The dVRK performs untangling of a cut elastic hairtie on a boxed, foam-padded surface to avoid end-effector damage and to prevent the cable from easily leaving the workspace. Given its small dimensions of 5 mm by 15 cm and its flexible material properties, we find this elastic cable to be conducive to manipulation with the dVRK. The workspace is equipped with an overhead Zivid OnePlus RGBD sensor which captures $1900 \times 1200$ RGBD image observations, although only the RGB channels are used in HULK, LOKI, and SPiDERMan inference. The dVRK is calibrated with a standard pixel-to-world camera transformation using the fiducial registration procedure described in~\cite{hwang2020efficiently}. The complete experimental setup has workspace bounds $\bm{w}_l,\bm{w}_r$. Each grasp is executed with a $30^{\circ}$ approach angle to avoid collisions caused by top down grasping.

\begin{table*}[!htbp]
\centering
 \begin{tabular}{|| c | c || c | c | c | c || c | c | c | c | c ||}
\hline
Tier & Policy & Success Rate & Node Deletion Actions & Recovery Actions & Total Actions & \multicolumn{5}{c||}{Failure Modes} \\ 
\hline
\multicolumn{6}{||c||}{}& A & B & C & D & E\\
\hline
\hline
1 & H & 6/12 & 6.5 & -- & 6.5 & 1 & 3 & 0 & 2 & 0 \\ 
\hline
1 & H+L & 6/12 & 5 & -- & 5 & 2 & 2 & 0 & 2 & 0 \\ 
\hline
1 & H+S & 6/12 & 5 & 3.5 & 8.5 &0 & 0 & 0 & 6 & 0\\ 
\hline
1 & H+L+S & \textbf{8/12} & 6.5 & 1 & 8 & 1& 1 & 2 & 0 & 0 \\
\hline
\hline
2 & H & 4/12 & 5.5 & -- & 5.5 & 1 & 6 & 0 & 1 & 0\\
\hline
2 & H+L & 7/12 & 5 & -- & 5 & 0 & 4 & 0 & 1 & 0 \\
\hline
2 & H+S & 8/12 & 5 & 0.5 & 7.5 & 2 & 0 & 0 & 1 & 1 \\ 
\hline
2 & H+L+S & \textbf{9/12} & 4 & 2 & 6 & 1 & 0 & 1 & 0 & 1 \\ 
\hline
\hline
3 & H & 0/12 & -- & -- & -- & 0 & 7 & 0 & 3 & 2\\
\hline
3 & H+L & 6/12 & 4 & -- & 4 & 1 & 5 & 0 & 0 & 0 \\
\hline
3 & H+S & 4/12 & 8.5 & 1 & 9.5 & 3 & 0 & 1 & 3 & 1 \\ 
\hline
3 & H+L+S & \textbf{8/12} & 5.5 & 0 & 6.5 & 1 & 0 & 1 & 2 & 0 \\ 
\hline
\hline
4 & H & 1/12 & 6 & -- & 6 & 3 & 6 & 0 & 1 & 1 \\ 
\hline
4 & H+L & 5/12 & 6 & -- & 6 & 3 & 4 & 0 & 0 & 0 \\
\hline
4 & H+S & 4/12 & 10.5 & 3.5 & 14 & 4 & 0 & 0 & 3 & 1 \\ 
\hline
4 & H+L+S & \textbf{10/12} & 9 & 3 & 12.5 & 2 & 0 & 0 & 0 & 0 \\
\hline
\hline
5 & H & 0/12 & -- & -- & -- & 1 & 10 & 0 & 1 & 0 \\
\hline
5 & H+L & 0/12 & -- & -- & -- & 1 & 9 & 0 & 1 & 1 \\ 
\hline
5 & H+S & 3/12 & 5 & 2 & 7& 1 & 0 & 2 & 6 & 0\\ 
\hline
5 & H+L+S & \textbf{6/12} & 8 & 2.5 &10.5& 2 & 0 & 0 & 4 & 0 \\ 
\hline
\end{tabular}

\caption{\textbf{Physical Results:}
\normalfont{
Success rate and efficiency (median number of actions per success) for untangling physical cables containing dense non-planar knots on the dVRK. We categorize initial cable configuration complexity into five tiers: (1) one semi-planar knot seen at train time, (2) two semi-planar knots both seen in training, (3) two semi-planar knots both seen in training and denser than those in (2), (4) one non-planar knot seen in training, and (5) one non-planar knot unseen in training. Untangling experiments are given a horizon of 5 node deletion actions to make progress. If the cable is equally or more dense after 5 untangling actions, we conclude that the configuration is pathological and terminate the trial. 
}}
\label{table:phys_exp_results}
\vspace{-0.85cm}
 \end{table*}

\subsection{Results}

Given the trained networks, we instantiate the proposed policies and baselines and run 12 trials of dVRK cable untangling for each method and each of the four tiers. At the beginning of each trial, a human supervisor manually ties a dense configuration and places it at $\bm{w}_c$. Then, the dVRK executes the untangling procedure in Algorithm~\ref{alg:untangle_alg} without intervention. A successful trial is defined as one that terminates such that the cable has at most one crossing and no knots. This definition accounts for the natural tendency of the cable to lie in single-crossing stable poses due to its elastic material properties. 
On an Nvidia GeForce RTX 2080, HULK keypoint inference takes 314 ms, LOKI offset and rotation inference takes 260 ms, and SPiDERMan density comparison takes 18 ms. Node Deletion and Reidemeister moves each take 10 s to execute, while Re-Posing and Wedge Recovery moves take 7 s and 15 s to execute respectively.

We report the untangling success rate and median number of Node Deletion moves, Recovery moves (Re-Posing and Wedged Recovery), and total actions per success in Table~\ref{table:phys_exp_results}. These results suggest that the combination of HULK, LOKI, and SPiDERMan is effective in performing cable untangling, and exhibits higher empirical success with fewer actions than methods that do not jointly leverage all 3 algorithms. 
Untangling failures are distributed evenly across initial configuration type for Tiers 1, 3, and 4. For Tier 2, all 3 failures are with the Overhand + Figure 8 initial configuration. For Tier 5, all the initial configurations have one failure each, with the exception of the Ashley-Stopper knot, which had 2 failures in 2 trials. 

We observe 5 failure modes: 
\begin{enumerate}[(A)]
    \item gripper collision due to poor keypoint and/or grasp rotation predictions in high-density configurations; 
    \item robot gripper jaws wedged between cable segments due to high configuration density and/or failed recovery actions; 
    \item premature termination due to density comparison false positives  in Equation~\eqref{eq:rotate_condition}; 
    \item  termination due to lack of untangling progress;
    \item the cable suddenly springing out of the reachable workspace where Re-Posing moves cannot grasp the cable due to poor keypoint predictions and the cable's elastic material properties. 
\end{enumerate}
Across policies that do not leverage SPiDERMan, the most common failure mode is the tendency of the gripper jaws to become wedged between cable segments (B). This failure mode is exacerbated with increasing density and non-planarity, though somewhat alleviated by LOKI's grasp refinement. 
We note a 22\% decrease in HULK's performance on identical initial configurations from experiments in this work compared to those reported in~\citet{grannen2020untangling}: 4/12 (33\% success) in Table~\ref{table:phys_exp_results}.2H versus 11/20 (55\% success) from~\cite{grannen2020untangling}. Six of the eight failures were wedged gripper failures (B), which we hypothesize result from the more frayed cable texture in this work compared to \citet{grannen2020untangling}. This may be a consequence of elastic cable overuse over many trials and can be seen in the cables of Figure~\ref{configs-fig}. HULK as reported in ~\cite{grannen2020untangling} is also trained exclusively on semiplanar configurations while the HULK baseline in this paper is trained on non-planar configurations, presenting a harder perceptual task. Lastly, we postulate that 12 trials of untangling experiments per tier may have high variance, resulting in large differences in performance rates even across similar experimental setups.

SPiDERMan's sensing of configuration density change over time can either mistakenly or correctly detect a lack of untangling progress, resulting in premature (C) or justified rollout termination (D), respectively. Gripper collisions (A) and the cable springing to workspace extremities (E) account for the remaining manipulation-induced errors. Failure (A) is an artifact of high-density configurations and mispredicted grasp locations and orientations, while (E) can cause SPiDERMan's recovery moves to reach robot joint limits, yielding an irrecoverable state.

\section{Conclusion}
We present LOKI and SPiDERMan, two algorithms that increase precision and recovery for robot untangling of dense, non-planar knots. When used in conjunction with HULK from prior work, LOKI and SPiDERMan anticipate, refine, and recover from errors due to HULK's coarse action planning. 
We experimentally evaluate the separate and collective impact of HULK, LOKI, and SPiDERMan on physical untangling of dense, non-planar knots across five difficulty tiers. 
We find that when combined, HULK, LOKI and SPiDERMan enable success rates of 72.8\% and 50\% on untangling knots of seen and unseen classes respectively and notably achieve a success rate of 66.7\% on untangling non-planar knots not considered in prior work. In future work, we will consider cables of varying visual and material properties and explore how depth sensing or other sensing modalities can improve performance. We will also consider other grasp parametrizations and explore how ideas from LOKI and SPiDERMan can be applied more broadly to other deformable manipulation tasks.

\section*{Acknowledgments}
\footnotesize

This research was performed at the AUTOLAB at UC Berkeley in affiliation with the Berkeley AI Research (BAIR) Lab,  the Real-Time Intelligent Secure Execution (RISE) Lab
and the CITRIS ``People and Robots" (CPAR) Initiative. Any opinions, findings, and conclusions or recommendations expressed in this material are those of the author(s) and do not necessarily reflect the views of the sponsors. The authors were supported in part by donations from Toyota Research Institute, Google, and by equipment grants from Intuitive Surgical. The da Vinci Research Kit is supported by the National Science Foundation, via the National Robotics Initiative (NRI), as part of the collaborative research project ``Software Framework for Research in
Semi-Autonomous Teleoperation" between The Johns Hopkins University (IIS 1637789), Worcester Polytechnic Institute (IIS 1637759), and the University of Washington (IIS 1637444). Ashwin Balakrishna is supported by an NSF GRFP. We thank our colleagues who provided helpful feedback and suggestions, especially Daniel Seita and Rishi Parikh.

\bibliographystyle{plainnat}
\bibliography{references}

\begin{thebibliography}{35}
\providecommand{\natexlab}[1]{#1}
\providecommand{\url}[1]{\texttt{#1}}
\expandafter\ifx\csname urlstyle\endcsname\relax
  \providecommand{\doi}[1]{doi: #1}\else
  \providecommand{\doi}{doi: \begingroup \urlstyle{rm}\Url}\fi

\bibitem[Bradski and Kaehler(2008)]{bradski2008learning}
Gary Bradski and Adrian Kaehler.
\newblock \emph{Learning OpenCV: Computer vision with the OpenCV library}.
\newblock O'Reilly Media, Inc., 2008.

\bibitem[Cappell(2010)]{cappell2010injury}
Mitchell~S Cappell.
\newblock Injury to endoscopic personnel from tripping over exposed cords,
  wires, and tubing in the endoscopy suite: a preventable cause of potentially
  severe workplace injury.
\newblock \emph{Digestive diseases and sciences}, 55\penalty0 (4):\penalty0
  947--951, 2010.

\bibitem[Chi and Berenson(2019)]{chi2019occlusion}
Cheng Chi and Dmitry Berenson.
\newblock Occlusion-robust deformable object tracking without physics
  simulation.
\newblock In \emph{2019 IEEE/RSJ Int. Conf. on Intelligent Robots and Systems
  (IROS)}, pages 6443--6450. IEEE, 2019.

\bibitem[Donald(1989)]{donald1989error}
Bruce~R Donald.
\newblock \emph{Error detection and recovery in robotics}, volume 336.
\newblock Springer, 1989.

\bibitem[Florence et~al.(2018)Florence, Manuelli, and
  Tedrake]{florence2018dense}
Peter~R Florence, Lucas Manuelli, and Russ Tedrake.
\newblock Dense object nets: Learning dense visual object descriptors by and
  for robotic manipulation.
\newblock In \emph{Conf. on Robot Learning (CoRL)}, 2018.

\bibitem[Ganapathi et~al.(2021)Ganapathi, Sundaresan, Thananjeyan, Balakrishna,
  Seita, Grannen, Hwang, Hoque, Gonzalez, Jamali,
  et~al.]{ganapathi2020learning}
Aditya Ganapathi, Priya Sundaresan, Brijen Thananjeyan, Ashwin Balakrishna,
  Daniel Seita, Jennifer Grannen, Minho Hwang, Ryan Hoque, Joseph~E Gonzalez,
  Nawid Jamali, et~al.
\newblock Learning to smooth and fold real fabric using dense object
  descriptors trained on synthetic color images.
\newblock In \emph{{Proc. {IEEE} Int. Conf. Robotics and Automation (ICRA)}},
  2021.

\bibitem[Grannen et~al.(2020)Grannen, Sundaresan, Thananjeyan, Ichnowski,
  Balakrishna, Hwang, Viswanath, Laskey, Gonzalez, and
  Goldberg]{grannen2020untangling}
Jennifer Grannen, Priya Sundaresan, Brijen Thananjeyan, Jeffrey Ichnowski,
  Ashwin Balakrishna, Minho Hwang, Vainavi Viswanath, Michael Laskey, Joseph~E
  Gonzalez, and Ken Goldberg.
\newblock Untangling dense knots by learning task-relevant keypoints.
\newblock In \emph{Conf. on Robot Learning (CoRL)}, 2020.

\bibitem[He et~al.(2016)He, Zhang, Ren, and Sun]{he2016identity}
Kaiming He, Xiangyu Zhang, Shaoqing Ren, and Jian Sun.
\newblock Identity mappings in deep residual networks.
\newblock In \emph{European conference on computer vision}, pages 630--645.
  Springer, 2016.

\bibitem[Henrich and W{\"o}rn(2012)]{henrich2012robot}
Dominik Henrich and Heinz W{\"o}rn.
\newblock \emph{Robot manipulation of deformable objects}.
\newblock Springer Science \& Business Media, 2012.

\bibitem[Hoque et~al.(2020)Hoque, Seita, Balakrishna, Ganapathi, Tanwani,
  Jamali, Yamane, Iba, and Goldberg]{hoque2020visuospatial}
Ryan Hoque, Daniel Seita, Ashwin Balakrishna, Aditya Ganapathi, Ajay~Kumar
  Tanwani, Nawid Jamali, Katsu Yamane, Soshi Iba, and Ken Goldberg.
\newblock Visuospatial foresight for multi-step, multi-task fabric
  manipulation.
\newblock In \emph{Proc. Robotics: Science and Systems (RSS)}, 2020.

\bibitem[Hwang et~al.(2020)Hwang, Thananjeyan, Paradis, Seita, Ichnowski, Fer,
  Low, and Goldberg]{hwang2020efficiently}
Minho Hwang, Brijen Thananjeyan, Samuel Paradis, Daniel Seita, Jeffrey
  Ichnowski, Danyal Fer, Thomas Low, and Ken Goldberg.
\newblock Efficiently calibrating cable-driven surgical robots with rgbd
  fiducial sensing and recurrent neural networks.
\newblock \emph{IEEE Robotics and Automation Letters}, 5\penalty0 (4):\penalty0
  5937--5944, 2020.

\bibitem[Jiang et~al.(2011)Jiang, Koo, Kikuchi, Konno, and
  Uchiyama]{jiang2011robotized}
Xin Jiang, Kyong-mo Koo, Kohei Kikuchi, Atsushi Konno, and Masaru Uchiyama.
\newblock Robotized assembly of a wire harness in a car production line.
\newblock \emph{Advanced Robotics}, 25\penalty0 (3-4):\penalty0 473--489, 2011.

\bibitem[Johannink et~al.(2019)Johannink, Bahl, Nair, Luo, Kumar, Loskyll,
  Ojea, Solowjow, and Levine]{johannink2019residual}
Tobias Johannink, Shikhar Bahl, Ashvin Nair, Jianlan Luo, Avinash Kumar,
  Matthias Loskyll, Juan~Aparicio Ojea, Eugen Solowjow, and Sergey Levine.
\newblock Residual reinforcement learning for robot control.
\newblock In \emph{2019 International Conference on Robotics and Automation
  (ICRA)}, pages 6023--6029. IEEE, 2019.

\bibitem[Kazanzides et~al.(2014)Kazanzides, Chen, Deguet, Fischer, Taylor, and
  DiMaio]{kazanzides-chen-etal-icra-2014}
Peter Kazanzides, Zihan Chen, Anton Deguet, Gregory~S. Fischer, Russell~H.
  Taylor, and Simon~P. DiMaio.
\newblock An open-source research kit for the da {V}inci surgical system.
\newblock In \emph{{Proc. {IEEE} Int. Conf. Robotics and Automation (ICRA)}},
  2014.

\bibitem[Lallemand(1998)]{lallemand1998blender}
Thomas Lallemand.
\newblock Blender, June~9 1998.
\newblock US Patent App. 29/074,220.

\bibitem[Lee et~al.(2020{\natexlab{a}})Lee, Florensa, Tremblay, Ratliff, Garg,
  Ramos, and Fox]{lee2020guided}
Michelle~A Lee, Carlos Florensa, Jonathan Tremblay, Nathan Ratliff, Animesh
  Garg, Fabio Ramos, and Dieter Fox.
\newblock Guided uncertainty-aware policy optimization: Combining learning and
  model-based strategies for sample-efficient policy learning.
\newblock In \emph{2020 IEEE International Conference on Robotics and
  Automation (ICRA)}, pages 7505--7512. IEEE, 2020{\natexlab{a}}.

\bibitem[Lee et~al.(2020{\natexlab{b}})Lee, Ward, Cosgun, Dasagi, Corke, and
  Leitner]{lee2020learning}
Robert Lee, Daniel Ward, Akansel Cosgun, Vibhavari Dasagi, Peter Corke, and
  Jurgen Leitner.
\newblock Learning arbitrary-goal fabric folding with one hour of real robot
  experience.
\newblock In \emph{Conf. on Robot Learning (CoRL)}, 2020{\natexlab{b}}.

\bibitem[Lin et~al.(2020)Lin, Wang, Olkin, and Held]{lin2020softgym}
Xingyu Lin, Yufei Wang, Jake Olkin, and David Held.
\newblock {SoftGym}: Benchmarking deep reinforcement learning for deformable
  object manipulation.
\newblock In \emph{Conf. on Robot Learning (CoRL)}, 2020.

\bibitem[Lui and Saxena(2013)]{lui2013tangled}
Wen~Hao Lui and Ashutosh Saxena.
\newblock Tangled: Learning to untangle ropes with {RGB-D} perception.
\newblock In \emph{2013 IEEE/RSJ Int. Conf. on Intelligent Robots and Systems},
  pages 837--844. IEEE, 2013.

\bibitem[Matas et~al.(2018)Matas, James, and Davison]{matas2018sim}
Jan Matas, Stephen James, and Andrew~J Davison.
\newblock Sim-to-real reinforcement learning for deformable object
  manipulation.
\newblock In \emph{Conf. on Robot Learning (CoRL)}, 2018.

\bibitem[Nair et~al.(2017)Nair, Chen, Agrawal, Isola, Abbeel, Malik, and
  Levine]{nair2017combining}
Ashvin Nair, Dian Chen, Pulkit Agrawal, Phillip Isola, Pieter Abbeel, Jitendra
  Malik, and Sergey Levine.
\newblock Combining self-supervised learning and imitation for vision-based
  rope manipulation.
\newblock In \emph{2017 IEEE Int. Conf. on Robotics and Automation (ICRA)},
  pages 2146--2153. IEEE, 2017.

\bibitem[Paradis et~al.(2021)Paradis, Hwang, Thananjeyan, Ichnowski, Seita,
  Fer, Low, Gonzalez, and Goldberg]{paradis2020intermittent}
Samuel Paradis, Minho Hwang, Brijen Thananjeyan, Jeffrey Ichnowski, Daniel
  Seita, Danyal Fer, Thomas Low, Joseph~E Gonzalez, and Ken Goldberg.
\newblock Intermittent visual servoing: Efficiently learning policies robust to
  instrument changes for high-precision surgical manipulation.
\newblock In \emph{{Proc. {IEEE} Int. Conf. Robotics and Automation (ICRA)}},
  2021.

\bibitem[S{\'a}nchez et~al.(2020)S{\'a}nchez, Wan, and
  Harada]{sanchez2020tethered}
Daniel S{\'a}nchez, Weiwei Wan, and Kensuke Harada.
\newblock Tethered tool manipulation planning with cable maneuvering.
\newblock \emph{IEEE Robotics and Automation Letters}, 5\penalty0 (2):\penalty0
  2777--2784, 2020.

\bibitem[Schmidt et~al.(2016)Schmidt, Newcombe, and Fox]{schmidt2016self}
Tanner Schmidt, Richard Newcombe, and Dieter Fox.
\newblock Self-supervised visual descriptor learning for dense correspondence.
\newblock \emph{IEEE Robotics and Automation Letters}, 2\penalty0 (2):\penalty0
  420--427, 2016.

\bibitem[Seita et~al.(2020)Seita, Ganapathi, Hoque, Hwang, Cen, Tanwani,
  Balakrishna, Thananjeyan, Ichnowski, Jamali, et~al.]{seita2019deep}
Daniel Seita, Aditya Ganapathi, Ryan Hoque, Minho Hwang, Edward Cen, Ajay~Kumar
  Tanwani, Ashwin Balakrishna, Brijen Thananjeyan, Jeffrey Ichnowski, Nawid
  Jamali, et~al.
\newblock Deep imitation learning of sequential fabric smoothing from an
  algorithmic supervisor.
\newblock In \emph{Proc. IEEE/RSJ Int. Conf. on Intelligent Robots and Systems
  (IROS)}, 2020.

\bibitem[Seita et~al.(2021)Seita, Florence, Tompson, Coumans, Sindhwani,
  Goldberg, and Zeng]{seita2020learning}
Daniel Seita, Pete Florence, Jonathan Tompson, Erwin Coumans, Vikas Sindhwani,
  Ken Goldberg, and Andy Zeng.
\newblock Learning to rearrange deformable cables, fabrics, and bags with
  goal-conditioned transporter networks.
\newblock In \emph{{Proc. {IEEE} Int. Conf. Robotics and Automation (ICRA)}},
  2021.

\bibitem[She et~al.(2020)She, Wang, Dong, Sunil, Rodriguez, and
  Adelson]{she2019cable}
Yu~She, Shaoxiong Wang, Siyuan Dong, Neha Sunil, Alberto Rodriguez, and Edward
  Adelson.
\newblock Cable manipulation with a tactile-reactive gripper.
\newblock In \emph{Proc. Robotics: Science and Systems (RSS)}, 2020.

\bibitem[Sundaresan et~al.(2020)Sundaresan, Grannen, Thananjeyan, Balakrishna,
  Laskey, Stone, Gonzalez, and Goldberg]{sundaresan2020learning}
Priya Sundaresan, Jennifer Grannen, Brijen Thananjeyan, Ashwin Balakrishna,
  Michael Laskey, Kevin Stone, Joseph~E Gonzalez, and Ken Goldberg.
\newblock Learning rope manipulation policies using dense object descriptors
  trained on synthetic depth data.
\newblock In \emph{{Proc. {IEEE} Int. Conf. Robotics and Automation (ICRA)}},
  2020.

\bibitem[Suzuki et~al.(1985)]{suzuki1985topological}
Satoshi Suzuki et~al.
\newblock Topological structural analysis of digitized binary images by border
  following.
\newblock \emph{Computer vision, graphics, and image processing}, 30\penalty0
  (1):\penalty0 32--46, 1985.

\bibitem[Thananjeyan et~al.(2021)Thananjeyan, Balakrishna, Nair, Luo,
  Srinivasan, Hwang, Gonzalez, Ibarz, Finn, and Goldberg]{recovery-rl}
Brijen Thananjeyan, Ashwin Balakrishna, Suraj Nair, Michael Luo, Krishnan
  Srinivasan, Minho Hwang, Joseph~E. Gonzalez, Julian Ibarz, Chelsea Finn, and
  Ken Goldberg.
\newblock Recovery {RL}: Safe reinforcement learning with learned recovery
  zones.
\newblock \emph{Robotics and Automation Letters}, 2021.

\bibitem[Van Den~Berg et~al.(2010)Van Den~Berg, Miller, Duckworth, Hu, Wan, Fu,
  Goldberg, and Abbeel]{van2010superhuman}
Jur Van Den~Berg, Stephen Miller, Daniel Duckworth, Humphrey Hu, Andrew Wan,
  Xiao-Yu Fu, Ken Goldberg, and Pieter Abbeel.
\newblock Superhuman performance of surgical tasks by robots using iterative
  learning from human-guided demonstrations.
\newblock In \emph{2010 IEEE International Conference on Robotics and
  Automation}, pages 2074--2081. IEEE, 2010.

\bibitem[Wang and Kroemer(2019)]{wang2019learning}
Austin~S Wang and Oliver Kroemer.
\newblock Learning robust manipulation strategies with multimodal state
  transition models and recovery heuristics.
\newblock In \emph{2019 International Conference on Robotics and Automation
  (ICRA)}, pages 1309--1315. IEEE, 2019.

\bibitem[Yan et~al.(2020{\natexlab{a}})Yan, Zhu, Jin, and Bohg]{yan2020self}
Mengyuan Yan, Yilin Zhu, Ning Jin, and Jeannette Bohg.
\newblock Self-supervised learning of state estimation for manipulating
  deformable linear objects.
\newblock \emph{IEEE Robotics and Automation Letters}, 5\penalty0 (2):\penalty0
  2372--2379, 2020{\natexlab{a}}.

\bibitem[Yan et~al.(2020{\natexlab{b}})Yan, Vangipuram, Abbeel, and
  Pinto]{yan2020learning}
Wilson Yan, Ashwin Vangipuram, Pieter Abbeel, and Lerrel Pinto.
\newblock Learning predictive representations for deformable objects using
  contrastive estimation.
\newblock In \emph{Conf. on Robot Learning (CoRL)}, 2020{\natexlab{b}}.

\bibitem[Zhang et~al.(2021)Zhang, Ichnowski, Seita, Wang, and
  Goldberg]{zhang2020robots}
Harry Zhang, Jeffrey Ichnowski, Daniel Seita, Jonathan Wang, and Ken Goldberg.
\newblock Robots of the lost arc: Learning to dynamically manipulate
  fixed-endpoint ropes and cables.
\newblock In \emph{{Proc. {IEEE} Int. Conf. Robotics and Automation (ICRA)}},
  2021.

\end{thebibliography}

\end{document}